\documentclass[runningheads]{llncs}
\usepackage{graphicx}
\usepackage{multirow}
\usepackage{algorithm} 
\usepackage{algpseudocode}
\usepackage{subcaption}
\usepackage{amsmath}
\usepackage{amssymb}
\usepackage{amsfonts}
\usepackage{marvosym}
\usepackage{rotating}
\usepackage{adjustbox}
\usepackage{enumitem}
\usepackage{booktabs}
\usepackage{threeparttable}
\usepackage{threeparttablex}
\usepackage{tabularx}
\usepackage{caption}
\usepackage{wrapfig}
\usepackage[table]{xcolor}

\begin{document}
%
% \title{A Spatio-temporal Semantics-based Abstract Modeling For Verify Deep Reinforcement Learning\thanks{Supported by organization x.}}
% \title{
% Advancing ICPS Decision-Making with Spatio-temporal Value Semantic Modeling in DRL}
\title{Spatio-temporal Value Semantics-based Abstraction for Dense Deep Reinforcement Learning}
%Spatio-temporal Semantic-based abstraction for Dense DRL?如果强调dense，需要在introduction及相关工作中进行介绍、说明，突出我们的motivation

%"Advancing ICPS Decision-Making with Spatio-temporal Value Semantic Modeling in DRL"
% "Simplifying State Space in Deep Reinforcement Learning: A Spatio-temporal Semantic Approach"
% "Enhancing Efficiency in ICPS through Abstract Modeling in Deep Reinforcement Learning"

\titlerunning{Abbreviated paper title}

\author{Jihui Nie\inst{1}
% \orcidID{0000-1111-2222-3333}
\and
Dehui Du\inst{1}\textsuperscript{(\Letter)} \and
Jiangnan Zhao\inst{1}}
\institute{Shanghai Key Laboratory of Trustworthy Computing, ECNU, Shanghai, China
\email{dhdu@sei.ecnu.edu.cn	}}

\maketitle              % typeset the header of the contribution
\begin{abstract}
% 全称， 空格， 引用, 空格
Intelligent Cyber-Physical Systems (ICPS) represent a specialized form of Cyber-Physical System (CPS) that incorporates intelligent components, notably Convolutional Neural Networks (CNNs) and Deep Reinforcement Learning (DRL), to undertake multifaceted tasks encompassing perception, decision-making, and control. The utilization of DRL for decision-making facilitates dynamic interaction with the environment, generating control actions aimed at maximizing cumulative rewards. Nevertheless, the inherent uncertainty of the operational environment and the intricate nature of ICPS necessitate exploration within complex and dynamic state spaces during the learning phase. DRL confronts challenges in terms of efficiency, generalization capabilities, and data scarcity during decision-making process. In response to these challenges, we propose an innovative abstract modeling approach grounded in spatial-temporal value semantics, capturing the evolution in the distribution of semantic value across time and space. A semantics-based abstraction is introduced to construct an abstract Markov Decision Process (MDP) for the DRL learning process. Furthermore, optimization techniques for abstraction are delineated, aiming to refine the abstract model and mitigate semantic gaps between abstract and concrete states. The efficacy of the abstract modeling is assessed through the evaluation and analysis of the abstract MDP model using PRISM. A series of experiments are conducted, involving diverse scenarios such as lane-keeping, adaptive cruise control, and intersection crossroad assistance, to demonstrate the effectiveness of our abstracting approach.
% The experimental results show that our approach is effective in terms of model simplicity, accuracy, and semantic equivalence. Our approach enables the semantic-preserving abstraction for DRL, which pave the way to formal analyze the learning process of machine learning.
\end{abstract}

% Abstract Model -> Abstract MDP

%重写。抛问题
\section{Introduction}
%一定要写清楚为什么要抽象，维度太高 决策困难 举例子
%强调Motivation

%1.基于模型的强化学习能够提高数据的有效性，然而现有的初始学习模型不能真实的反映现实场景，太过符号化的抽象，使得基于模型的训练的强化学习泛化性性能较差，当场景发生微小变化时，可能需要重新补充模型，进而重新进行强化学习的训练
 
The Cyber-Physical System (CPS) integrates computing, networking, and physical environments, expertly coordinated by computer and communication components with a joint monitoring mechanism~\cite{DBLP:journals/ijsi/ZhangDZZWZ21}. The evolution of the Intelligent Cyber-Physical System (ICPS) as a mainstream paradigm is marked by the integration of AI-enabled components such as controllers or sensors~\cite{DBLP:journals/ais/RadanlievRKSA21}. Notably, the utilization of Deep Reinforcement Learning (DRL) in decision-making, as emphasized by Brunke et al.~\cite{brunke2022safe}, is particularly promising due to its innate ability for dynamic interaction within the environment.

However, the efficacy of DRL encounters a formidable challenge in the form of an expansive state space, leading to prolonged algorithmic convergence and heightened complexities in the formal verification of the learning process. Furthermore, the inherent black-box nature of DRL poses challenges when applied to safety-critical ICPS scenarios, such as Autonomous Driving Systems (ADSs) and robots. To surmount the challenge posed by the vast state space in ICPS tasks, strategic compression becomes imperative. Dense DRL is a 
promising field to address these issues\cite{feng2023dense}. 
% 用引用别人的话进行分析
An effective abstraction modeling in this context involves leveraging prior knowledge for generalization from concrete to abstract states~\cite{li2006towards}. 
% This aligns with the contemporary categorization of DRL abstract modeling, encompassing three distinct strategies.

The existing abstraction modeling methods can be roughly divided into three categories. In \textit{the first category}, the focus lies on abstracting similar states, thereby addressing the challenge of sparse rewards in DRL~\cite{dockhorn2023state,nashed2022selecting,li2022phasic}. Strategic games, such as Go~\cite{DBLP:journals/nature/SilverHMGSDSAPL16}, utilize hierarchical organization based on the significance of empty points on the board's corners and edges. This approach, inspired by consistent patterns in Go, allows the amalgamation of game elements into a unified abstract state, offering a potential solution for the slow convergence issue. Real-time strategy games~\cite{DBLP:journals/nature/MnihKSRVBGRFOPB15} similarly represent states through collections of game elements and positions. \textit{The second category} involves temporal abstraction, extending decision-making over time~\cite{kulkarni2016hierarchical,bacon2017option,jiang2019language}. This concept extends to natural language processing, where units of state are often identified as words or phrases through corpus analysis~\cite{DBLP:journals/tacl/BojanowskiGJM17,DBLP:conf/emnlp/PenningtonSM14}. This offers a promising avenue for effectively compressing the state space, thereby facilitating improved algorithmic convergence in DRL. In the realm of ICPS, \textit{the third category} focuses on state-action abstraction, a method extensively applied in addressing the intricate challenges unique to these systems. Notably, empirical evidence establishes a polynomial relationship between the number of ICPS samples and the state space's size~\cite{abel2022theory}, emphasizing the nuanced nature of the challenge. For instance, the operation of ADS unfolds within open and dynamic environments characterized by inherent complexity and unpredictability. This complexity manifests in two key aspects. Firstly, the expansive state space explored by DRL incurs substantial exploration costs due to its high dimensions, contributing to discernible scalability limitations, as extensively discussed in existing literature~\cite{DBLP:journals/toit/YangXPML18}. Secondly, the state-action abstraction approach, while effective in enhancing abstract efficiency and addressing gaps in the abstract model, tends to overlook the consistency of the constructed model in the formal method, leading to potential distortions in the model's representation.

In the delineated classification, it becomes evident that traditional approaches to abstraction typically concentrate on minimizing model size while concurrently preserving model accuracy. However, when confronted with the complexities of ICPS, characterized by abundant randomness and uncertainty in the state space, these models frequently fall short of fulfilling the requisites for formal verification. Establishing a verifiable abstract model for ICPS is an imperative undertaking, underscoring the necessity to meticulously tailor the granularity of abstraction to the specific requirements and intricacies of the application context. Striking a delicate balance between mitigating state explosion and sustaining optimal model performance is crucial, as expounded in the work by Schmidt et al.~\cite{DBLP:conf/itsc/SchmidtBPEM22}.
%我们是属于状态抽象！ 给点明！！！

%重新描述图片！

% In this paper, we introduce a pioneering abstraction approach employing \textit{spatio-temporal value semantics} for dense DRL, as detailed in Fig.~\ref{fig:Overview}. This approach meticulously integrates spatial and temporal dimensions, state valuations, and pivotal determinants of action selection to construct an abstract MDP. The essence of this approach lies in its capacity to condense the state and action spaces through a tailored spatio-temporal value metric. To evaluate the fidelity and efficiency of abstract MDP, we introduce specific indices and utilize PRISM for semantic equivalence verification. Furthermore, we implement this abstract model within an online setting, devising a joint strategy through abstract MDP-Guided DRL. The practicality and validity of our approach are corroborated through empirical experiments, demonstrating the effectiveness of our abstract modeling approach.

% 图文不匹配
\begin{figure}[ht]
\setlength{\abovecaptionskip}{0.cm}
\centering\includegraphics[width=\columnwidth]{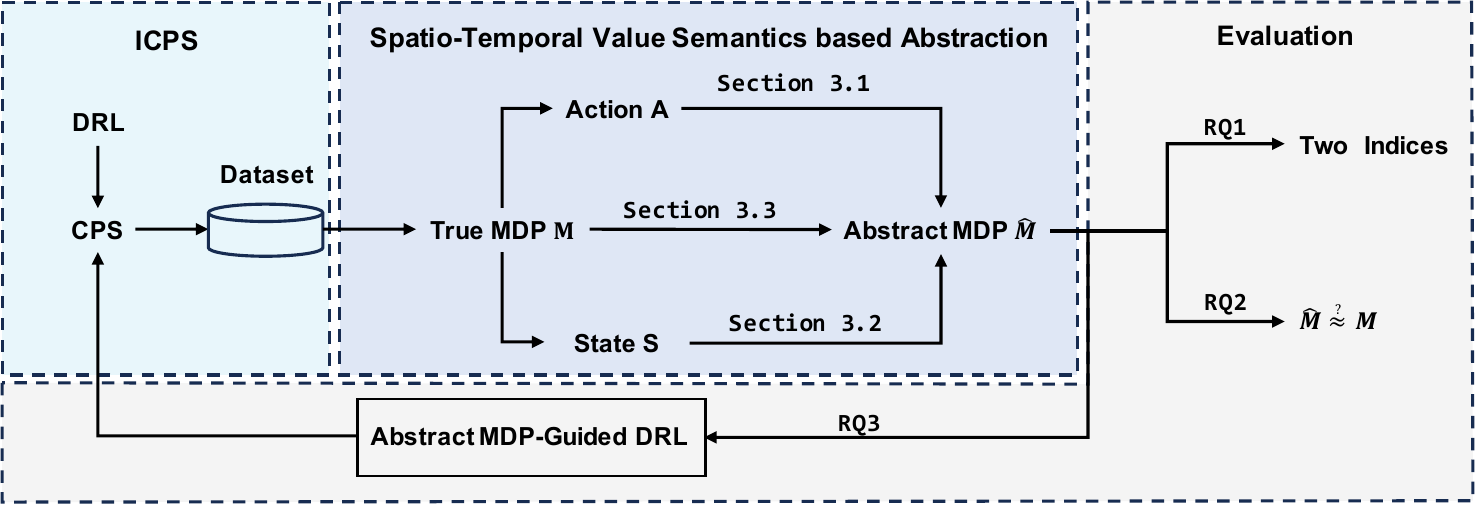}
\caption{The Framework of Our Approach}
\label{fig:Overview}
\end{figure}

The main contributions in our work include:
\begin{enumerate}
    \item We propose the spatio-temporal value metric to measure the similarity between states. It helps to make semantic-preserving abstraction effectively.
    \item We propose a novel abstraction modeling approach to construct the abstract MDP model based on spatio-temporal value metric. Moreover, we optimize the abstract model to make the abstract state space closer to the original state space.
    \item We use compression ratio and mean absolute error to measure the accuracy of the model. We innovatively use PRISM to check for semantic gaps between abstract and real models. We use abstract MDP to build a dense DRL framework to generate joint policies in online environments.
\end{enumerate}

\textbf{Paper organization}. The remainder of the paper is organized as follows. In Section~\ref{section:2}, we provide a review of pertinent background and preliminary information. In Section~\ref{Section:3}, we present the overarching framework of our approach. In Section~\ref{Section:4}, we conduct an assessment of our approach through several case studies related to ADS. Finally, in Section~\ref{Section:5}, we review related work, and our conclusions are summarized in Section~\ref{Section:6}.

\section{Preliminaries}\label{section:2}
% In the following, we review the relevant background related with DRL and existing abstraction techniques.
% 增加Euclidean metric，multi-step metric
\subsection{Deep Reinforcement Learning}
DRL is an amalgamation of reinforcement learning (RL) and deep learning, training agents to achieve goals in simulated or real environments through a sequence of decision-making actions. By learning through continuous trial and error, DRL enables agents to explore environments and discover optimal strategies. At each time step, a DRL agent receives state information from the environment, which is used as input to a deep neural network (DNN). Subsequently, the agent selects an action from a set of possible actions and receives real-time rewards, aiming to maximize cumulative rewards. The process of DRL can be formalized using MDPs.

\subsection{Markov Decision Process}

\begin{definition}[Markov Decision Process]\label{MDP} 
A Markov Decision Process is denoted by a tuple $M = (S, s_0, A, R, P, \gamma)
$, where $S$ represents a finite non-empty state set, $s_0 \in S$ denotes the initial system state, $A$ stands for a finite non-empty action set, $P: S \times A \times S \rightarrow [0,1]$ is the transition probability function such that for $s \in S$ and $a \in A$, $\sum_{s' \in S} P(s, a, s') = 1$. $R: S \times A \rightarrow \mathbb{R}$ represents the reward assigned to the current state-action pair and $\gamma \in (0, 1)$ is the discount factor. The discount factor $\gamma$ determines the importance of immediate rewards relative to future rewards. A policy for the MDP denoted as $\pi: S \rightarrow A$, maps states to actions.\end{definition}

The MDP $M$ describes the evolution of the initial system state over discrete time steps. In DRL, the interaction between the policy $\pi$ and the environment, resulting in state transitions and immediate rewards forms the foundation of the learning process. The evaluation and optimization of the policy involve state value function and action value function.

\begin{definition}[State Value Function $V(s)$]\label{V(s)} The state value function $V(s)$ represents the expected return achievable under a policy $\pi$ from state $s$. The Bellman expectation equation for $V(s)$ in recursive form is:

\begin{equation}
    V(s) = \sum_{a \in A} \pi(a|s) [R(s,a) + \gamma \sum_{s' \in S} P(s'|s,a) V(s')].
\end{equation}

\end{definition}

\begin{definition}[Action Value Function $Q(s, a)$]\label{Q(s,a)} The action value function $Q(s, a)$ represents the expected return achievable from taking action $a$ in state $s$ and following policy $\pi$. The Bellman expectation equation for $Q(s, a)$ in recursive form is:
\begin{small}
    \begin{equation}
Q(s, a) = R(s,a) + \gamma \sum_{s' \in S} P(s'|s,a) \sum_{a' \in A} \pi(a'|s') Q(s', a').
\end{equation}
\end{small}

\end{definition}

With the rapid development in the field of DRL, many online, model-free learning algorithms have been proposed to fulfill various requirements~\cite{brunke2022safe}, such as Deep Deterministic Policy Gradient (DDPG), Twin Delayed Deep Deterministic Policy Gradient (TD3), Actor-Critic (A2C), Proximal Policy Optimization (PPO), and more. The use of DRL controllers in place of traditional controllers holds great promise in large-scale systems with complex dynamics.
% To model the DRL learning process with abstracting technology, we propose a new abstract model called the abstract Markov decision model.
\subsection{Abstract Markov Decision Process}

\begin{definition}[Abstract Markov Decision Process]\label{Abstract_Markov_Model}
     Let $M = (S, s_0, A, R,\\ P, \gamma)$ denotes the true MDP, and $\hat{M} = (\hat{S}, \hat{s}_0, \hat{A}, \hat{P}, \hat{R}, \gamma, \hat{\pi})$ denotes the abstract MDP. $\Phi: S \rightarrow \hat{S}$ is the state abstraction function, where $\Phi(s) \in \hat{S}$ represents the abstract state, and its inverse image is denoted as $\Phi^{-1}(\hat{s})$, where $\hat{s} \in \hat{S}$. $\hat{S}$ is the basic state set corresponding to the abstraction function $\Phi$. $\Psi: A \rightarrow \hat{A}$ is the action abstraction function, where $\Psi(a) \in \hat{A}$ represents the abstract action, and its inverse image is denoted as $\Psi^{-1}(\hat{a})$, where $\hat{a} \in \hat{A}$. $\hat{A}$ is the basic action set corresponding to the abstraction function $\Psi$.
\end{definition}

State transition and reward functions are defined as follows:
\begin{equation}
\hat{R}(\hat{s}, \hat{a}) = \sum_{s \in \Phi^{-1}(\hat{s}),\, a \in \Psi^{-1}(\hat{a})} w(s) v(a) R(s, a),
\end{equation}

\begin{equation}
\hat{P}_{\hat{s}\hat{s}'}^{\hat{a}} = \sum_{s \in \Phi^{-1}(\hat{s}),\, a \in \Psi^{-1}(\hat{a})} \sum_{s' \in \Phi^{-1}(\hat{s'})} w(s) v(a) P_{ss'}^a,
\end{equation}
where $w: S \rightarrow [0,1]$, $\sum_{s \in \Phi^{-1}(\hat{s})} w(s) = 1$ represents the weigh function for state, and $v: A \rightarrow [0,1]$, $\sum_{a \in \Psi^{-1}(\hat{a})} v(a) = 1$ represents the weight function for action. $\hat{R}(\hat{s}, \hat{a})$ represents the immediate reward of transitioning from abstract state $\hat{s}$ to $\hat{s}'$ after taking the abstract action $\hat{a}$. $\hat{P}_{\hat{s}\hat{s}'}^{\hat{a}}$ represents the probability of transitioning from abstract state $\hat{s}$ to $\hat{s}'$ after taking the abstract action $\hat{a}$. The abstract policy $\hat{\pi}$ is generated based on the abstract MDP.
% and is consistent with the original policy $\pi$.

\subsection{PRISM}
PRISM serves as an open-source probabilistic model checker for the formal modeling and analyzing of probabilistic systems~\cite{kwiatkowska2004probabilistic,parker2003implementation}. Widely applied in diverse application domains, PRISM has been instrumental in analyzing systems ranging from communication and multimedia protocols to randomized distributed algorithms, security protocols, biological systems, and beyond. PRISM is proficient in constructing and scrutinizing a variety of probabilistic models, encompassing: Discrete-time and continuous-time Markov chains (DTMCs and CTMCs), MDPs and probabilistic automata (PA). %Probabilistic timed automata (PTAs), partially observable MDPs and PTAs (POMDPs and POPTAs). Interval Markov chains and MDPs (IDTMCs and IMDPs). 
Additionally, PRISM supports extensions of these models that incorporate cost and reward considerations. It facilitates automated analysis of a broad spectrum of quantitative properties inherent in these models. For instance, users can inquire about the probability of a system shutdown within 4 hours due to a failure, the worst-case probability of a protocol terminating in error across all potential initial configurations, the anticipated size of a message queue after 30 minutes, or the worst-case expected time for an algorithm to conclude. The property specification language integrates temporal logic such as PCTL, CSL, LTL, and PCTL*, alongside extensions for quantitative specifications, costs, and rewards.

\section{Spatio-temporal Value Semantics based Abstraction for DRL}\label{Section:3}

\begin{figure}[ht]
  \centering
    \includegraphics[width=\linewidth, height=2in, keepaspectratio]{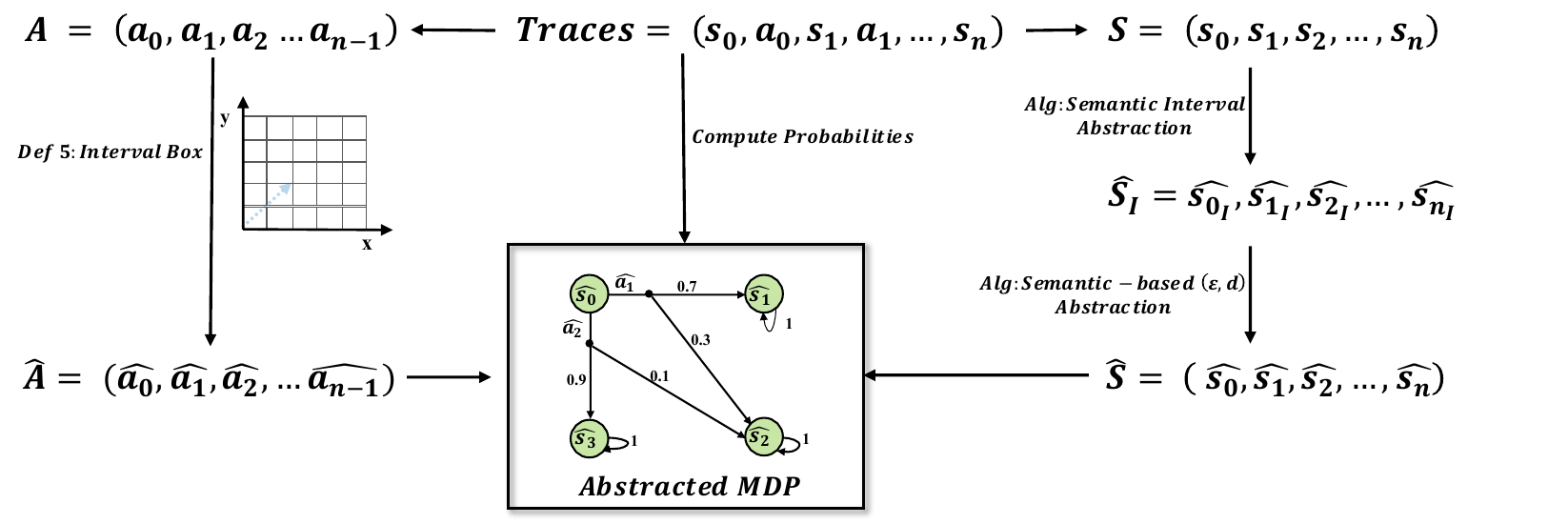}
    \captionof{figure}{Semantic-based Abstraction for DRL}
    \label{fig:abstract_framework}
\end{figure}

For ensuring the safety and reliability of ICPS, the design and optimization of controllers play a pivotal role. However, the hybrid behavior of the system and the uncertainties in the environment contribute to a vast state space for ICPS, rendering the design and optimization of controllers using DRL a complex task. 
%复杂度不能解决，降低
To address this complexity, we propose an abstraction modeling approach based on spatio-temporal value semantics, aiming to efficiently abstract the state space and actions of ICPS, thereby facilitating the optimization of controller design through DRL. The key aspect of state abstraction revolves around ensuring semantic consistency, which involves measuring the similarity between different states and determining their belongingness to the same abstract state. To achieve this, we introduce a novel measurement approach termed spatio-temporal value semantics. Fig.~\ref{fig:abstract_framework} illustrates the process of abstracting DRL based on spatio-temporal value semantics.

\subsection{Action Abstraction}\label{Action abstraction}

The controller in ICPS usually selects a specific value in a continuous range as the specific value of the action. Taking ADS as an example, the acceleration range is usually in $[-8m/s^2,3m/s^2]$, and the accelerator and brake of the vehicle make the vehicle achieve the expected acceleration. However, the triggering action of MDP is usually a concrete value in this range. It is obviously unrealistic to realize the construction of an MDP for this infinite number of concrete actions in ICPS.

To address this issue, we propose a technique for abstracting continuous action spaces. The fundamental idea is to finely segment the continuous action space so that the abstract action is analogous to the action on the true MDP transition, with analogous successor states and rewards gained.
Inspired by~\cite{DBLP:conf/cav/JinTZWZ22}, we introduce the interval action box as follows:
\begin{definition}[Interval Box]\label{def:interval_box}
     For a continuous action space $A$ of dimension $d$, each variable in each dimension has its own effective range, i.e., the variable $a_i$ in the $i-th$ dimension ($i \in [0…d]$) is in the range $[l_i, u_i]$. The interval box method divides this range uniformly into unit intervals $I_i = [l_i, u_i]/g_i$, which is a partition of the continuous action space $A$. For an action $a$, based on the interval box abstraction, the action $\hat{a} = [k_1, k_2, …, k_d]$ of the abstract action space $\hat{A}$ is obtained, where $k_i = a_i / g_i$, $g_i$ is the abstraction granularity of the $i-th$ dimension.
\end{definition}
According to Definition~\ref{def:interval_box}, the abstraction granularity must be adjusted in the abstract MDP to maintain the successor state and reward gained by performing the abstract action close to the successor state and the reward obtained by executing the actual action in the true MDP. Each dimension's level of granularity needs to be specified by the specific environment.
Different cases call for varying levels of granularity since fine granularity is closer to the original action, while too small granularity would amplify data jitters inaccuracy.

% \begin{figure}[ht]
% \setlength{\abovecaptionskip}{0.cm}
% \centering\includegraphics[width=0.8\columnwidth]{picture/Action_Abstraction.pdf}
% \caption{\textbf{\textit{Action Abstraction}}}
% \label{action_abstraction}
% \end{figure}
 
\subsection{Semantics-based State
Abstraction}\label{State_abstraction}
Existing state abstraction approaches struggle to handle the high dimensionality and continuous state space of ICPS. %Specifically, strategy-independent and Q-value-independent abstraction approaches fail to accurately capture the behavioral characteristics of controllers based on DRL.
While model-irrelevance abstraction is an ideal solution in theory, it is challenging to implement in practice. To address these issues, we propose a semantic-based abstraction approach to introduce value information, spatio-temporal information, and probability information into the abstraction process, which enables the evaluation of the similarity between states. We first introduce the semantic-based abstraction model.
\begin{definition}[Semantic-based Abstraction MDP]\label{def:Semantic-Based Abstraction Model} An abstraction model is denoted by a tuple $(\hat{S}, \hat{A}, \hat{s}_0, \eta, \Theta)$, where $\hat{S}$ is the set of abstract states, $\hat{A}$ is the set of abstract actions, $\eta: \hat{S} \times \hat{A} \times \hat{S} \rightarrow [0, 1]$ represents the transition distribution, $\hat{s}_0$ is the initial state set, and $\Theta$ represents the mapping to the semantic space.
\end{definition}

It is worth noting that the core of semantic abstraction lies in measuring the similarity of states in abstract MDP. The similar states must satisfy two conditions: (1) the available action sets of similar states should be similar, and (2) the multi-step state transition models and multi-step rewards of similar states should be similar.
\subsubsection{Semantic Interval Abstraction }

To satisfy the specified conditions, we implement \textit{Semantic Interval Abstraction}. Taking Adaptive Cruise Control (ACC) as an example, where the goal is to maintain a safe distance from the lead vehicle. The concrete state $s$ of the ego vehicle, represented as a multidimensional vector $(v, acc, x, y, distance, \ldots)$, includes parameters like vehicle speed $v$, acceleration $acc$, spatial coordinates ($x$, $y$), and relative distance.

Based on existing causal discovery algorithms, we employ PC (Peter-Clark) algorithm~\cite{spirtes1997reply} and FCI (Fast Causal Inference) algorithm~\cite{spirtes2013causal} to construct causal graphs on the autonomous driving dataset. The union of these two causal graphs is considered as the final causal graph. By examining the causal relationships within the graph, we identify the relationships among different dimensions of the state. Through abstraction based on these causal relationships, we determine the semantic values.

Due to the continuous nature of each dimension, the combination of specific dimensions results in a vast state space. In the context of ICPS, information from the state vector exhibits approximate similarity within short time intervals. Thus, the concrete state $s$ undergoes semantic abstraction, yielding a condensed representation $d = (\text{$rel_{\text{velocity}}$}, \text{$rel_{\text{angle}}$}, \text{$rel_{\text{distance}}$},\ldots)$, where \text{$rel_{\text{velocity}}$}, \text{$rel_{angle}$}, and \text{$rel_{distance}$} represent relative velocity, relative angle, and relative distance, respectively. In summary, semantic interval abstraction maps multiple dimensions of a concrete state into several semantic values, achieving abstraction of state dimensions.

Since the dimensions of the semantic value $\theta$ may have different scales, the data needs to be normalized to a uniform scale. Therefore, we divide the $J$-dimensional space into $\prod_{j=1}^JK_j$ segments, with each dimension having $K_j$ intervals i.e.
$d_i^j = [l_i^j, u_i^j]$. In this context, $d_{i}^j$ represents the $i$-th interval on the $j$-th dimension, $l_i^j$ and $u_i^j$ are the lower and upper bounds of this interval. Consequently, the spatial partitioning problem is transformed into an optimization problem, specifically formulated as follows:
\begin{small}
    \begin{align}\label{eq6}
    &\begin{cases}
        \max \left(u_{i}^{j}-l_{i}^{j}\right) \\
        \text{s.t.}\begin{aligned}
            &d_{\text{MIN}}^{j} \le u_{i}^{j}-l_{i}^{j} \le d_{\text{MAX}}^{j} \\
            &\left|\hat{s_{i}^{j}}\right| \ge n_{\text{MIN}}^{j} \\
            &\text{MEAN}\left\{\hat{\theta_{s}^{j}} - E\left[\hat{\theta_{s}^{j}}\right]\right\} < e_{\text{MEAN}}^{j} \\
            &\text{MAX}\left(\hat{\theta_{s}^{j}} - E\left[\hat{\theta_{s}^{j}}\right]\right) < e_{\text{MAX}}^{j}
        \end{aligned}
    \end{cases}
\end{align}
\end{small}where $d_{\text{MIN}}^j$ and $d_{\text{MAX}}^j$ are the minimum and maximum lengths of intervals on the $j$-th semantic dimension. $\hat{s}_{i}^j = \{s|\theta_{s}^j\in d_{i}^j\}$ represents the set of concrete states with semantic values $\theta_{s}^j$ falling within the interval $d_{i}^j$. $n_{\text{MIN}}^j$ is the minimum number of concrete states in the $j$-th dimension interval, and $e_{\text{MEAN}}^j$ and $e_{\text{MAX}}^j$ are predefined expected error and maximum error for abstract on the $j$-th dimension. Eq.~\ref{eq6} ensures that each interval contains enough concrete states while maintaining low abstract errors.

% 算法1
%内容写清楚
%for 维度
\begin{algorithm}[h]
  \caption{Semantic Interval Abstraction Algorithm For State}
  \label{alg:interval_abstraction}
  \begin{algorithmic}[1]
  \Require Concrete state set $S$, Semantic value set $\theta$, Maximum interval lengths $d_{\text{MAX}}$, Minimum interval lengths $d_{\text{MIN}}$, Minimum number of concrete states in intervals $n_{\text{MIN}}$, Expected error $e_{\text{MEAN}}$, Maximum error $e_{\text{MAX}}$, $REDUCTION\_LEVEL$ $r_d$
  \Ensure Intervalized abstract space $\hat{S_{I}}$
  \State \texttt{refined $\leftarrow$ False, $\hat{S_{I}} \leftarrow \emptyset$} \hspace{5mm} \/// Initialize abstraction process
  \While {\texttt{refined = False}} \hspace{5mm} \/// Iterative refinement 
    \For {\texttt{$j\in\{1,...,J\}$}} \hspace{5mm} \/// Partition state set based on Eq.~\ref{eq6}
      \State \texttt{$d_{j}\leftarrow PARTITION(S,\theta_{j},d_{\text{MAX}},d_{\text{MIN}},n_{\text{MIN}})$}
    \EndFor
    \State \texttt{$D\leftarrow d_{1} ,...,d_{J}$} \hspace{5mm} \/  // Form set of intervalized dimensions
    \State \texttt{$\hat{S_{I}}\leftarrow STATE\_MAPPING(D)$} \/   // Map intervals to abstract space
    \State \texttt{$e_{mean}, e_{max}\leftarrow COMPUTE\_ERROR(\hat{S_{I}} ,S)$} \hspace{5mm}\/  // Compute current error \label{line:compute-error}
    \State \texttt{$r_{\text{cur}}\leftarrow COMPUTE\_REDUCTION\_LEVEL(\hat{S_{I}} ,S)$} \hspace{5mm}\/// Compute reduction level \label{line:compute-reduction-level}
    \If{\texttt{$e_{\text{mean}}>e_{\text{MEAN}}\ \ or\ \  e_{\text{max}}>e_{\text{MAX}}\ \ or\ \ r_{\text{cur}} > r_d$} } 
      \State \texttt{update $d_{\text{MAX}},d_{\text{MIN}},n_{\text{MIN}}$} \hspace{5mm}\/// Update interval parameters if needed
    \Else
      \State \texttt{refined $\leftarrow$ True}\hspace{5mm} \/// End refinement if conditions are met
    \EndIf
  \EndWhile
  \State \texttt{Return $\hat{S_{I}}$}\hspace{5mm} \/// Return intervalized abstract space
  \end{algorithmic}
\end{algorithm}

% 算法1
Alg.~\ref{alg:interval_abstraction} has been devised to orchestrate the transformation of a given concrete state set $S$ into an intervalized abstract space denoted as $\hat{S_{I}}$. This transformation takes into account semantic values and adheres to specific constraints. The essential parameters guiding this process encompass the semantic value set $\theta$, the maximum and minimum interval lengths $d_{\text{MAX}}$ and $d_{\text{MIN}}$, the minimum number of concrete states in intervals $n_{\text{MIN}}$, the expected error threshold $e_{\text{MEAN}}$, and the limit for reduction level $r_d$.

% The algorithm commences by initializing the refinement flag $refined$ to \texttt{False} and establishing an empty abstract space $\hat{S_{I}}$ (Line \ref{line:init-refined}). Subsequently, it iteratively partitions the concrete state set $S$ along each dimension based on semantic values. These partitions are created through the resolution of Eq.~\ref{eq6}, yielding corresponding intervals $d_j$(Line \ref{line:partition}) and forming the set $D$ (Line \ref{line:form-set-D}). The intervalized dimensions are then mapped onto the abstract space $\hat{S_{I}}$ using the \texttt{STATE\_MAPPING} function (Line \ref{line:state-mapping}). 
% Following this mapping, the algorithm proceeds to compute the current error $e_{\text{cur}}$ and the reduction level $r_{\text{cur}}$ (Lines \ref{line:compute-error}-\ref{line:compute-reduction-level}). 
% Line \ref{line:compute-error}, $e_{\text{cur}}$ is defined as: 
% \begin{small}
%     \begin{equation}
%     \text{MEAN}(\theta_{\hat{s_I}}) - E[\theta_s].
%     \end{equation}
% \end{small}
% \[\text{MEAN}(\theta_{\hat{s_I}}) - E[\theta_s].
% \]

The term REDUCTION\_LEVEL~\cite{song2023mathtt} serves as an indicator of the state compression ratio. The optimization iterations are designed to terminate when the abstract effect meets the predefined requirements of $r_{\text{d}}$. A reasonable range is to make the number of compressed states 10\% to 30\% of the original number of states. 

\subsubsection{Semantic-based ($\varepsilon$,d)-Abstraction} While the semantic interval abstraction can ensure model accuracy, abstraction simplicity is related to the granularity of abstraction. To address this, we propose $(\varepsilon,d) -abstraction$ based on spatio-temporal value semantics.

% To enhance the extraction of semantic information from real-world scenarios, we introduce spatio-temporal value semantics as a representation of concrete states.

\begin{definition}[Spatio-temporal Value Semantics]
    \label{def:Spatio-temporal_Value_Semantics}
     For a concrete state $s \in S$ and spatio-temporal value semantics $\theta \in \mathbb{R}^n$, $\theta = \Theta\{V(s), Q(s,a), R(s,a),\\ P(s,s'), \ldots\}$. Here, $\theta$ represents the semantic value, and the function $\Theta: S \rightarrow \theta$ maps states to their corresponding semantic values. The multidimensional mapping $\Theta\{V(s), Q(s, a), R(s, a), P(s, s'), \ldots\}$ captures various aspects of the state $s$, including the state value function $V(s)$, action value function $Q(s, a)$, reward function $R(s,a)$, and transition function $P(s, s')$, among others. The semantic mapping $\Theta$ translates state and environmental information into the semantic space, thus depicting the spatio-temporal value of the state.
\end{definition}

Spatio-temporal value semantics encapsulate state evolution information across time and space, reflecting the state's distribution and evolution. It considers factors like state value function and transition function, providing insights into both current and future states. Using spatio-temporal value semantics, we assess semantic equivalence between abstract states. When the semantic distance is below a specified threshold, states are considered equivalent, simplifying model abstraction with enhanced precision.

\begin{definition}[($\varepsilon$,d)-Abstraction]\label{def:ed-abstraction}
   ($\varepsilon$,d)-Abstraction is denoted by a mapping $\Phi_{\varepsilon,d}: S \rightarrow \hat{S}$ that satisfies the following condition:
   \begin{small}
       \begin{equation}
       d(s_1,s_2) \leq \varepsilon \quad \forall \hat{s}\in \hat{S}, \quad s_1, s_2\in \Phi_{\varepsilon,d}^{-1}(\hat{s}).
    \end{equation}
   \end{small}
    $\Phi: S\rightarrow \hat{S}$ represents the abstract mapping function that maps the original state space $S$ into an abstract state space $\hat{S}$. The mapping function $\Phi$ transforms a true MDP into an abstract MDP. Let $Pow(S)$ denote the power set of $S$, and $\Phi^{-1}: \hat{S} \rightarrow Pow(S)$ represents the inverse mapping function. The core of state abstraction is measuring the similarity of state abstractions and performing nearest-neighbor abstractions based on state similarity. $d$ represents the state distance metric, and $\varepsilon$ is the abstraction threshold.
\end{definition} 

Grounded in the state value functions (Eq.\ref{V(s)}) and action value functions (Eq.\ref{Q(s,a)}) in MDP, when two states possess akin transition models and rewards, their expected cumulative rewards will be similar. This observation provides a simplification approach for Semantic-based ($\varepsilon$,d)-Abstraction: the reward function and transition probabilities can compose the \textit{Spatio-temporal Value Metric} of that state. Thus, in the process of abstraction based on value semantics, we aim to maintain the optimal value function of the abstract MDP as close as possible to the true MDP, ensuring semantic equivalence.

\begin{definition}[Spatio-temporal Value Metric] \label{def:SpatiotemporalValueMetric}
    For $\forall  s_1, s_2 \in S$,
    \begin{small}
        \begin{equation}
        \begin{aligned}
            d(s_1,s_2) & = d(\theta_{s_1},\theta_{s_2}) \triangleq \max_{\hat{a} \in \hat{A}(s_1) \cap \hat{A}(s_2)} \big\{c_R \lvert R(s_1,\hat{a})-R(s_2,\hat{a})\rvert \\
            & \quad +c_Pd_P(P(\cdot|s_1,\hat{a}),P(\cdot|s_2,\hat{a}))\big\}+c_DD[\hat{A}(s_1),\hat{A}(s_2)] \\
            & \quad +c_TD[s_1,s_2],
        \end{aligned}
    \end{equation}
    \end{small}
\end{definition}
where $c_R$ and $c_P$ are positive constants, $d_P$ represents a measure function of the similarity between two probability distributions. $c_R$ and $c_P$ are weights for measuring rewards and transition probability density functions, respectively. $c_D$ and $c_T$ are sufficiently large positive constants, and $\hat{A}(s_1)$ is considered equivalent to $\hat{A}(s_2)$ if $c_DD[\hat{A}(s_1),\hat{A}(s_2)] \leq \epsilon$. The spatio-temporal value metric satisfies mutual simulatability and uniqueness, i.e., $d(s_1,s_2)=0$ implies $s_1=s_2$.

% This metric serves to evaluate present and future system states, capturing the impact of high-dimensional state spaces through semantics-based abstraction. Understanding system behavior is crucial for assessing current operational state quality and defining performance metrics, akin to specifying the reward function in RL. Spatio-temporal value semantics is applied in the context of ($\varepsilon$,d)-abstraction~\cite{guo2021state}.

\begin{algorithm}[h]
  % \caption{Semantic-based ($\varepsilon$,d)-Abstraction for States}
  \caption{Semantic-based ($\varepsilon$,d)-Abstraction Algorithm}
  \label{alg:ed-abstraction}
  \begin{algorithmic}[1]
  \Require Intervalized abstract space $\hat{S_{I}} = \{D_1, D_2, \ldots, D_n\}$, Optimal state number determination function $K$, Spatiotemporal Value Metric $d$
  \Ensure Semantic-based ($\varepsilon$,d)-Abstraction space $\hat{S}$, Abstraction Model $\Phi$
  \State Number of clusters $k \leftarrow$ $K(\hat{S_{I}})$ \label{alg:ed-abstraction-line1}
  \State Initialize cluster centroids randomly: $c_1, c_2, \ldots, c_k \leftarrow$ random points from $\hat{S_{I}}$ \label{alg:ed-abstraction-line2}
  \Repeat \label{alg:ed-abstraction-line3}
      \For{$i = 1$ to $n$} \label{alg:ed-abstraction-line4}
        \State Assign each data point $D_i$ to the nearest centroid: \label{alg:ed-abstraction-line5} \\
        \hspace*{2em} $c_{j(i)} = d(D_i, c_k)$ \label{alg:ed-abstraction-line6}
      \EndFor \label{alg:ed-abstraction-line7}

      \For{$k = 1$ to $k$} \label{alg:ed-abstraction-line8}
        \State Update each centroid as the mean of the assigned data points: \label{alg:ed-abstraction-line9} \\
        \hspace*{2em} $c_k = \frac{1}{\lvert \{i : j(i) = k\} \rvert} \sum_{i : j(i) = k} D_i$ \label{alg:ed-abstraction-line10}
      \EndFor \label{alg:ed-abstraction-line11}
    \Until{Convergence} \label{alg:ed-abstraction-line12}
  \State \textbf{Return} {$\hat{S}$, $\Phi${$c_1, c_2, \ldots, c_K$}} \label{alg:ed-abstraction-line13}
  \end{algorithmic}
\end{algorithm}

The spatio-temporal value metric is employed in the Semantic-based ($\varepsilon$,d)-Abstraction algorithm, as delineated in Alg.~\ref{alg:ed-abstraction}. The procedure outlined in Alg.~\ref{alg:ed-abstraction} takes the intervalized abstract space $\hat{S_{I}}$ as its input. Key steps in the algorithmic execution involve the determination of the number of clusters based on the optimal state number determination function $K$ (Line \ref{alg:ed-abstraction-line1}), initializing cluster centroids randomly (Line \ref{alg:ed-abstraction-line2}), and iteratively assigning data points to the nearest centroids while updating centroids until convergence (Lines \ref{alg:ed-abstraction-line3}-\ref{alg:ed-abstraction-line12}). The ultimate outcome comprises the Semantic-based ($\varepsilon$,d)-Abstraction space $\hat{S}$ and the Abstraction Model $\Phi$, where the centroids $c_1, c_2, \ldots, c_K$ encapsulate the final representation of the abstract space (Line \ref{alg:ed-abstraction-line13}).

The time complexity of Alg.~\ref{alg:ed-abstraction} is determined by its iterative clustering, which involves data point assignments to centroids and centroid updates until convergence. The assignment step has a complexity of $O(n \cdot k \cdot m)$, where $n$ is the number of data points, $k$ the number of clusters, and $m$ the dimensionality. Centroid updates have a complexity of $O(k \cdot m)$. The algorithm's effectiveness is influenced by initial centroid positions and the fixed number of clusters, requiring adjustments based on dataset characteristics for optimal semantic-based ($\varepsilon$,d)-abstraction.

\subsection{Construction of Abstract MDP}\label{State_refinement}
In this subsection, we introduce a comprehensive methodology for constructing an MDP, essential for formulating decision-making frameworks in stochastic environments. Fig.~\ref{fig:abstract_framework} illustrates the procedural details of the approach. Initiated by gathering detailed trajectory data comprising observed states and actions over time, this approach tackles the inherent complexity and high dimensionality of the raw data. Central to our approach is the systematic abstraction of this data, necessary for effective MDP modeling. The methodology encompasses two primary abstraction processes: action abstraction through interval box and state abstraction through spatio-temporal value semantics. Action abstraction involves discretizing the continuous and diverse real-world actions into distinct intervals, each representing a group of similar actions. This process simplifies the action space, enhancing tractability and computational feasibility. Simultaneously, state abstraction condenses the state space by Alg.~\ref{alg:interval_abstraction} and Alg.~\ref{alg:ed-abstraction}, thereby capturing their essential characteristics. These abstractions are pivotal in reducing complexity, allowing for more efficient computation and analysis.

The next critical step in our methodology is the statistical computation of transition probabilities, derived from the frequency of state transitions in the trajectory data. These probabilities reflect the likelihood of moving from one state to another given a specific action, mirroring the dynamics of real-world scenarios. When calculating transition probabilities, we employ Hoeffding's inequality to reduce errors, enhancing the accuracy of our probability estimates. After abstracting states and actions and incorporating these probabilities, we formulate the initial MDP model. This model may include a reward system based on state transitions. However, recognizing that the initial model may not fully align with real-world contexts, we engage in iterative refinement. This involves adjusting abstractions, recalculating probabilities, and redefining states and actions to enhance the model's empirical alignment with observed data.

% In conclusion, our methodology offers a detailed approach to constructing MDPs, beginning with high-dimensional real-world data and methodically abstracting it into a computationally efficient model. This model undergoes continual refinement and optimization, aligning it with specific objectives like reward maximization or cost minimization. The resultant optimized MDP is versatile, and applicable across various decision-making, predictive analytics, and complex dynamic system analysis contexts.

\section{Implementation and Evaluation}\label{Section:4}
% In this section, we evaluate the performance and effectiveness of our approach systematically to answer some research questions.
\subsection{Case Study}
We conduct experiments in three representative ADS scenarios, encompassing diverse driving environments with varying control specifications.

\textbf{Lane Keeping Assist (LKA)} is an advanced driving assistance module~\cite{leurent2018environment} crucial for automated driving. LKA evaluates the lateral offset $d_{\text{t}}$ and relative yaw angle $\theta_{\text{t}}$ to adjust the front wheel steering angle $\theta_{\text{steer}}$. Its objective is to minimize lateral deviation and yaw angle, aligning them close to zero, and ensuring the vehicle stays within the lane.

\textbf{Adaptive Cruise Control (ACC)} is an intelligent module~\cite{leurent2018environment} adjusting the car's speed based on the distance from the preceding vehicle. It manages the vehicle's acceleration $a_{\text{ego}}$ to maintain a safe relative distance $d_{\text{rel}}$ greater than $d_{\text{safe}}$. ACC targets the user-set cruise speed $v_{\text{set}}$, adapting to the preceding vehicle's movement controlled by $a_{\text{lead}}$. The safe distance dynamically adjusts according to the relative velocity.

\textbf{Intersection Crossroad Assistance (ICA)} enhances safety in complex intersections~\cite{leurent2018environment}, integrating LKA and ACC features. ICA determines optimal speed and direction, demonstrating randomness for adaptive control and versatility for flexible adaptation. It navigates intersecting roads, one intelligent vehicle, and multiple environmental vehicles, aiming to traverse the intersection successfully with left, straight, or right turns while avoiding deviations or collisions.

\subsection{Research Questions}
To assess the effectiveness of the abstract model based on spatio-temporal value semantics, we investigate the following research questions:

\textbf{Research Question 1 (RQ1):} How does the performance of the abstract model, grounded in spatio-temporal value semantics, fare in terms of both simplicity and accuracy?

% In the context of industrial-grade ICPS with large state and action spaces, a concise and accurate abstract model can help reduce problem complexity and enhance interpretability for further processing. However, there is a trade-off between simplicity and accuracy during the abstraction process, as an overly simplified abstract model may result in accuracy loss.

\textbf{Research Question 2 (RQ2):} Does the abstract MDP model exhibit decision-making performance that approximates that of the true model? Moreover, is there a semantic equivalence between the abstract and true models?

% **Research Question 2 (RQ2):** To what extent can abstract models, verified for semantic equivalence using PRISM, act as guiding mechanisms within Deep Reinforcement Learning (DRL)? Can they effectively address challenges related to suboptimal data utilization and limited generalization, thereby accelerating the training process?
% The question arises of whether using an abstract model for decision-making can approximate or even outperform the true MDP. It is primarily because, in practical applications, the true MDP may be challenging to use directly due to a large or overly complex state space. In such cases, if the abstract model can provide similar or better decision effectiveness, it greatly enhances its practical value.

%在线进行决策，解决数据稀缺、泛化性差、低效等问题
%抽象模型是否能对强化学习的学习过程起到指导性作用，解决数据利用率低，泛化性差的问题，从而加速训练？
\textbf{Research Question 3 (RQ3):} Can abstract models effectively guide the learning process in DRL, specifically addressing issues of low data utilization and poor generalization, consequently leading to accelerated training

\subsection{Experiment Setup}
% This section describes the experimental settings for the experimental cases, including detailed parameters for data collection and abstract model construction.
\subsubsection{Metrics for Comparison}

\textit{Euclidean Metric}
is used to measure the straight-line distance between two states in Euclidean space.
For states \(s_1(p_1, p_2, ..., p_n)\) and \(s_2(q_1, q_2, ..., q_n)\) in $S$, the Euclidean metric is defined as:
\begin{small}
    \begin{equation}
    d(s_1, s_2) = \sqrt{\sum_{i=1}^{n}(q_i - p_i)^2}.
    \end{equation}
\end{small}

%Guo在【16】中提出什么方法，我们受启发，沿袭使用
\textit{Multi-Step Metric:}
For any $s_1, s_2 \in S$, the multi-step metric is defined as:

\begin{small}
    \begin{equation}
        \begin{aligned}
        & d_{\mathrm{M}}\left(s_1, s_2\right) \triangleq \max _{o \in O\left(s_1\right) \cap O\left(s_2\right)}\left\{c_R\left|R\left(s_1, a\right)-R\left(s_2, o\right)\right|\right. \\
        & \left.\quad+c_P d_p\left(P\left(\cdot \mid s_1, o\right), P\left(\cdot \mid s_2, o\right)\right)\right\} \\
        & \quad+c_\mathbb{D} \mathbb{D}[A(s_1),A(s_2)],
        \end{aligned}
    \end{equation}
\end{small}where $c_R, c_P$, and $c_{\mathbb{H}}$ are positive constants. $A(s)$ is the set of available actions for each state $s$. 
The function $\mathbb{D}[x, y]=0$ if $x=y$, and 1 otherwise. 
$c_{\mathbb{D}}$ is a sufficiently large constant such that $d_{\mathrm{M}}\left(s_1, s_2\right) \leq \epsilon$ implies that $A\left(s_1\right)$ equals $A\left(s_2\right)$~\cite{guo2021state}.

\subsubsection{DRL Setup}
During the data collection phase, we employ curiosity-driven TD3 with Random Network Distillation (RND)~\cite{DBLP:conf/iclr/BurdaESK19} to derive control strategies for LKA and ACC. Additionally, curiosity-driven DQN is utilized to generate control strategies for ICA, exploring the case environment and gathering system trajectories.

In each scenario, we simulate the curiosity-driven RL controller 1000 times to accumulate experience. This experience is then partitioned into a modeling set and a validation set in an $8:2$ ratio. The former is utilized for constructing the abstract MDP, while the latter is employed to scrutinize the semantic gaps between the abstract MDP and the concrete MDP.
The hyperparameter configurations for the deep learning networks in the three cases are detailed in Table~\ref{tab:algorithm_parameters}.
\begin{small}
    \begin{table}[H]
  \caption{Hyper-parameters of DRL}
  \label{tab:algorithm_parameters}
  \centering
  \begin{tabular}{|c|c|c|c|c|c|c|c|c|}
    \hline
    \multirow{2}{*}{Case Study} & \multirow{2}{*}{Algorithm} &\multirow{2}{*}{Activate function} &\multirow{2}{*}{Size} & \multicolumn{2}{c|}{Learning rate} & \multirow{2}{*}{$\gamma$} & \multirow{2}{*}{$\epsilon$} & \multirow{2}{*}{Soft tau} \\ \cline{5-6}
                          &                            &                          &                         & Critic & Actor   &                          &                        &   \\ \hline
    LKA                   & TD3                        & ReLU & $2 \times 128$                         & 1.00E-03  & 1.00E-03 & 0.95                   & /                        & 1.00E-02                  \\ \hline
    ACC                   & TD3                        & ReLU & $2 \times 128$                         & 1.00E-04  & 1.00E-04 & 0.95                   & /                        & 1.00E-02                  \\ \hline
    ICA                   & DQN                        & ReLU & $2 \times 128$                         & \multicolumn{2}{c|}{2.00E-03}   & 0.98                   & 0.01                     & /                         \\ \hline
  \end{tabular}
\end{table}
\end{small}

The rewards for LKA, ACC, and ICA are specified as follows:
\begin{small}
    \begin{align*}
        \text{LKA Reward:} & \quad r^{t} = 1-d_t^2 - \cos^2\theta_t ,\\
        \text{ACC Reward:} & \quad r^{t} = 0.05 \cdot v_t ,\\
        \text{ICA Reward:} & \quad r^{t} = 0.05 \cdot v_t - 0.0005 \cdot d^{t}_{g}.
    \end{align*}
    \begin{itemize}
      \item \( r^{t} \) is the reward at time step \( t \).
      \item \( d_t \) represents the lateral offset of the vehicle's current position from the lane center. A smaller \( d_t \) leads to a higher reward.
      \item \( \theta_t \) represents the angle between the current direction of the vehicle and the lane center direction. A smaller \( \theta_t \) leads to a higher reward.
      \item \( v_t \) represents the velocity of the vehicle at time step \( t \).
      \item \( d^{t}_{g} \) represents the distance from the target \(g\) at time \( t \).
    \end{itemize}
\end{small}

\subsubsection{Abstraction Setup}
The hyper-parameters for the abstraction process are shown in Table~\ref{tab:abstract-para}. We define the average semantic error, denoted as \(e_{\text{MEAN}}\), to be 0.005, and the maximum semantic error, denoted as \(e_{\text{MAX}}\), to be 0.01. These values represent 0.25\% of the overall value range. Simultaneously, we set the REDUCTION\_LEVEL (\(r_d\)) to 0.5\%. The parameter $k$ is determined through the elbow method, average silhouette method, and gap statistic method~\cite{kassambara2017practical}. We conduct a comparative analysis of abstractions using the traditional Euclidean distance method, the Multi-Step distance abstraction based on the state-of-the-art method proposed by Guo et al.~\cite{guo2021state}, and the abstraction employing the spatio-temporal value metric.
\begin{small}
    \begin{table}[H]
  \caption{Hyper-parameters of Abstraction Modeling
  }
  \label{tab:abstract-para}
  \centering
  \begin{tabular}{cccccc}
    \hline
    Case study                & Semantics & $d_{\text{min}}$ & $d_{\text{max}}$ & $n_{\text{min}}$\\ \hline
    LKA                       & $d_t$      & 0.001     &   0.005   &  1\%      \\ 
                             & $\theta _t$& 0.001     &   0.005   &  0.1\%    \\ \hline
    ACC                       & $v_t$      & 0.010     &   0.050   &  0.5\%    \\ \hline
    ICA                       & $v_t$      & 0.010     &   0.050   &  0.5\%    \\ 
                             & $d_g^t$    & 0.001     &   0.005   &  1\%     \\ \hline 
  \end{tabular}
\end{table}
\end{small}
\subsubsection{Two Indices}
To address \textbf{RQ1}, we evaluated the abstraction models with a focus on simplicity and accuracy using two indices: Compression Ratio (\textit{CR}) and Mean Absolute Error (\textit{MAE}). The formulas for \textit{CR} and \textit{MAE} are as follows:
\begin{small}
    \begin{equation}
        \begin{aligned}
            \textit{CR} =\frac{{\left\lvert \hat{S} \right\rvert}}{{\left\lvert S \right\rvert}},\\ 
        \end{aligned}
\end{equation}
\end{small}

\begin{small}
    \begin{equation}
        \begin{aligned}
            \textit{MAE} = \frac{1}{n} \sum_{i=1}^{n} \text{MEAN}\left\lvert y - \hat{y} \right\rvert, 
        \end{aligned}
    \end{equation}
\end{small}where $\left\lvert \hat{S} \right\rvert$ represents the number of abstract states, $\left\lvert S \right\rvert$ represents the number of original concrete states, $y$ is the prediction output of the abstract model, $\hat{y}$ is the output of the true model, and $\text{MEAN}\left\lvert y - \hat{y} \right\rvert$ measures the average deviation from the reference value. $n$ denotes the number of abstract states generated in a single experiment. \textit{CR} assesses the simplicity of the abstraction model, indicating the quality of the abstraction effect, while \textit{MAE} reveals the accuracy of the abstraction model in preserving the original semantic information. 

% 这些东西 有了这些东西可以做什么！！！ 应该放在方法里面～
\subsubsection{Abstract MDP-guided Training for DRL}
% 为了探究抽象MDP是否对DRL具有指导作用，我们修改三个ICPS的环境变量，以期望MDP指导的DRL学习过程表现更加安全，在DRL中引入了抽象MDP对动作选择的影响，目标是尽可能的加速DRL收敛过程，构建安全，可泛化的联合策略，使用抽象MDP的动作输出作为神经网络（NN）输出的影响变量。具体而言，agent结合抽象MDP的外在动作,获得联合动作，可以表示为:
To investigate the potential guiding impact of an abstract MDP on DRL, we systematically modify the environmental variables within three ICPS. We aim to foster a more secure DRL learning process under the guidance of the abstract MDP. We introduce the influence of the abstract MDP on action selection within the DRL framework, with the overarching objective of expediting the convergence of DRL and establishing a joint strategy that is both safe and generalized.

The influence of the abstract MDP on action selection is incorporated into the DRL by leveraging the abstract MDP's action output as a variable influencing the output of the neural network (NN). This integration is expressed mathematically as:
\begin{small}
    \begin{equation}
    \begin{aligned}
        \textit{a} = \alpha \cdot a_{NN} + \beta \cdot a_{MDP},
    \end{aligned}
    \end{equation}
\end{small}where \(\textit{a}\) represents the joint action output, \(a_{NN}\) denotes the action output from the NN, and \(a_{MDP}\) signifies the action output from the abstract MDP. The coefficients \(\alpha\) and \(\beta\) are set to 0.5 based on existing empirical values, reflecting the joint contributions of the NN and the abstract MDP in the synthesized action. 
% This equation illustrates the synthesis of external actions from the abstract MDP with those from the NN, providing a mechanism to optimize the joint strategy in the pursuit of accelerated and secure DRL convergence. 

% 在本文的实验环境下，\alpha，\beta 我们均设置为 0.5.

%图1展示了原始轨迹分布和经过两阶段抽象后的状态分布
% \subsubsection{Abstraction Effects Based on spatio-temporal  Value Semantics}
\subsection{Experimental Results and Analysis}
\textbf{RQ1: The Simplicity and Accuracy of Abstract MDP} Our investigation into the performance of abstract models, guided by spatio-temporal value semantics, is revealed through a combination of visual and quantitative data analyses. Fig.~\ref{fig:trace_distribution} offers a striking visual narrative of this examination. Fig.~\ref{fig:ACC_Abstract_Traces} shows we transform detailed raw data into a coherent, abstract grid. Within this grid, distinct hues represent individual abstract states, effectively simplifying the complex state space without compromising its comprehensive nature. This approach exemplifies how abstract modeling can achieve a harmonious blend of simplicity in design with precision in data representation.

The spatio-temporal value metric's efficacy in capturing the essence of the state space without overcomplication is quantitatively reinforced by Tab.~\ref{tb:ComparisonAnalysisofDifferentMetrics}. Here, the \textit{CR} and \textit{MAE} serve as the principal indices for evaluation. The \textit{CR} index, reflecting the proportionate reduction in state-space size, alongside the \textit{MAE} index, indicating the average error magnitude, collectively substantiates the spatio-temporal value metric's superior performance across various ICPS. Notably, in the ACC scenario, the spatio-temporal approach achieves \textit{CR}s of 10.12\% and 13.02\%, underscoring a substantial simplification of the model. Concurrently, the approach's consistently lower \textit{MAE} values across all scenarios, compared to Euclidean and Multi-Step approaches, highlight its enhanced accuracy.

\textbf{Response to RQ1:}  Spatio-temporal value metric simplifies the state space without sacrificing semantic accuracy, ensuring the abstract model's effectiveness in algorithmic analysis and decision-making. The model's simplicity is achieved without a concomitant increase in error, demonstrating an optimal balance between the two desired attributes of computational efficiency and semantic precision. These findings articulate the spatio-temporal value metric's pivotal role in generating abstract models that are not only operationally feasible but also semantically representative of the real-world systems they aim to emulate.

\textbf{RQ2: Semantic Equivalence in Abstract MDP} 
% delves into the comparative analysis of abstract MDP models and true MDPs, particularly focusing on their impact on decision-making efficacy and the assurance of semantic equivalence.
To address \textbf{RQ2}, we leveraged the PRISM model checker for encoding the abstraction model, ensuring the retention of critical state information such as rewards, transition probabilities, and metrics pertaining to lane adherence and collision incidents. We crafted specific properties that resonate with rewards and safety considerations, thereby facilitating a comprehensive evaluation of the extent to which the abstract model parallels the true MDP in semantic terms. This approach enabled a thorough exploration of the decision-making implications inherent in the abstract model, shedding light on its strengths and constraints.

In Tab~\ref{tab:Semantic_error}, we presented the encoded abstraction models for diverse scenarios using PRISM, emphasizing properties that encapsulate semantic information. Through PRISM, the semantic equivalence between the abstract and the true models was quantitatively assessed. For instance, in scenarios such as a four-way intersection, various properties like $Rmin=?[C<=60]$ (minimum expected cumulative reward within 60 time steps), $Pmax=?[F<=60; isOutOfLane=1]$ (maximum probability of lane departure within 60 steps), and $Pmax=?[F<=60; isCrashed=1]$ (maximum probability of collision involvement within 60 time steps) were analyzed to measure decision-making effects.

Our analysis in the sphere of MDP modeling, particularly through semantic gap analysis via PRISM, highlights the pronounced superiority of the spatio-temporal value approach over the Euclidean and Multi-Step approaches. This conclusion is drawn from a systematic juxtaposition across varied scenarios like LKA, ACC, and ICA. The spatio-temporal value approach consistently exhibited lower discrepancies in predicting minimum expected cumulative rewards and maximum probabilities of specific events, thereby indicating a higher fidelity in mimicking real-world dynamics. This aspect is especially pronounced in intricate scenarios like ACC and ICA, where the approach's accuracy in capturing semantic nuances suggests its enhanced capability in semantic property representation. The proficiency of the spatio-temporal value approach in bridging semantic gaps accentuates its robustness and reliability as a tool in stochastic decision-making frameworks, where fidelity to real-world conditions is paramount.

\textbf{Response to RQ2:} The findings corroborate that abstraction based on spatio-temporal values not only ensures semantic alignment with the true model but also offers substantial insights for refining training strategies. This aligns with the overarching goal of achieving a harmonious balance between model abstraction and real-world decision-making accuracy.

%压缩率 全称
\begin{small}
    \begin{table}[htb]
        \centering
        \caption{Comparison Analysis of Different Metrics}
        \label{tb:ComparisonAnalysisofDifferentMetrics}
        \resizebox{\textwidth}{!}{%
        {\fontsize{8}{10}\selectfont % Set the font size to 8pt with 10pt leading
        \begin{tabular}{c|c|c|c|c|c|c|c|c|c}
        \toprule
        \multirow{2}{*}{Case Study} & \multirow{2}{*}{Number of States} & \multirow{2}{*}{\scriptsize $k$} & \multirow{2}{*}{\scriptsize Metric} & \multirow{2}{*}{\scriptsize Abstract States$^{1^{st}}$} & \multirow{2}{*}{\scriptsize Abstract States$^{2^{nd}}$} & \multicolumn{2}{c}{ \textit{CR}} & \multicolumn{2}{c}{\textit{MAE}} \\
        \cmidrule(lr){7-8} \cmidrule(lr){9-10}
        & & & & & & \scriptsize \textit{CR}$^{1^{st}}$ & \scriptsize \textit{CR}$^{2^{nd}}$ & \scriptsize \textit{MAE}$^{1^{st}}$ & \scriptsize \textit{MAE}$^{2^{nd}}$ \\
        \midrule
            &     &      & Euclidean                  & \multirow{12}{*}{2931} & 2000         & \multirow{12}{*}{24.38\%} & 16.87\%          &  \multirow{12}{*}{17.08} & 191.696 \\
            &     &Elbow & Multi-Step                 &                        & 1930         &                           & 16.28\%          &                          & 85.008 \\
            &     &      &\cellcolor{gray!25} \textbf{Spatio-temporal}   &                        &\cellcolor{gray!25} \textbf{1930}&                           &\cellcolor{gray!25} \textbf{16.28\%} &                          &\cellcolor{gray!25} \textbf{43.43}  \\    
                 \cline{3-4} \cline{6-6}  \cline{8-8} \cline{10-10}
            &     &      & Euclidean                  &                        & 2170         &                           & 18.30\%          &                          & 169.578 \\
            &     &Silhouette & Multi-Step            &                        & 1200         &                           & 10.12\%          &                          & 107.043 \\
        ACC &11858&      &\cellcolor{gray!25} \textbf{Spatio-temporal}   &                        &\cellcolor{gray!25} \textbf{1200}&                           &\cellcolor{gray!25} \textbf{10.12\%} &                          &\cellcolor{gray!25} \textbf{46.796} \\ 
                  \cline{3-4} \cline{6-6}  \cline{8-8} \cline{10-10}
            &     &      & Euclidean                  &                        & 2210         &                           & 18.64\%          &                          & 168.003 \\
            &     & Gap   & Multi-Step                &                        & 1260         &                           & 10.63\%          &                          & 109.632 \\
            &     &      &\cellcolor{gray!25} \textbf{Spatio-temporal}   &                        &\cellcolor{gray!25} \textbf{1260}&                           &\cellcolor{gray!25} \textbf{10.63\%} &                          &\cellcolor{gray!25} \textbf{48.189} \\  \cline{3-4} \cline{6-6}  \cline{8-10}
            &     &       & Euclidean                 &                        & 1765         &                           & 14.88\%          &                          & 199.788 \\
            &     &Canopy & Multi-Step                &                        & 1544         &                           & 13.02\%          &                          & 84.018 \\
            &     &       &\cellcolor{gray!25} \textbf{Spatio-temporal}  &                        &\cellcolor{gray!25} \textbf{1544}&                           &\cellcolor{gray!25} \textbf{13.02\%} &                          &\cellcolor{gray!25} \textbf{46.419} \\ \cline{1-10}

            &     &      & Euclidean                  & \multirow{12}{*}{3804} & 2400         &\multirow{12}{*}{26.93\%}  & 16.99\%          &  \multirow{12}{*}{6.892} & 51.344 \\
            &     &Elbow & Multi-Step                 &                        & 1870         &                           & 13.24\%          &                          & 28.151 \\
            &     &      &\cellcolor{gray!25} \textbf{Spatio-temporal}   &                        &\cellcolor{gray!25} \textbf{1870} &                           &\cellcolor{gray!25} \textbf{13.24\%} &                          &\cellcolor{gray!25} \textbf{11.698}  \\    
                 \cline{3-4} \cline{6-6}  \cline{8-8} \cline{10-10}
            &     &           & Euclidean             &                        & 2760         &                           & 19.54\%          &                          & 28.229 \\
            &     &Silhouette & Multi-Step            &                        & 1530         &                           & 10.83\%          &                          & 37.016 \\
        LKA &14124&      &\cellcolor{gray!25} \textbf{Spatio-temporal}   &                        &\cellcolor{gray!25} \textbf{1530} &                           &\cellcolor{gray!25} \textbf{10.83\%}  &                          &\cellcolor{gray!25}  \textbf{16.347} \\ 
                 \cline{3-4} \cline{6-6}  \cline{8-8} \cline{10-10}
            &     &           & Euclidean             &                        & 2880         &                           & 20.39\%          &                          & 19.134 \\
            &     & Gap       & Multi-Step            &                        & 2060         &                           & 14.59\%          &                          & 37.2 \\
            &     &       &\cellcolor{gray!25} \textbf{Spatio-temporal}  &                        &\cellcolor{gray!25} \textbf{2060} &                           &\cellcolor{gray!25} \textbf{14.59\%} &                          &\cellcolor{gray!25} \textbf{16.533} \\   \cline{3-4} \cline{6-6}  \cline{8-8} \cline{10-10}
            &     &           & Euclidean             &                        & 2995         &                           & 21.21\%          &                          & 51.327 \\
            &     &Canopy     & Multi-Step            &                        & 2572         &                           & 18.21\%          &                          & 19.469 \\
            &     &       &\cellcolor{gray!25} \textbf{Spatio-temporal}  &                        &\cellcolor{gray!25}\textbf{2572} &                           &\cellcolor{gray!25} \textbf{18.21\%} &                          &\cellcolor{gray!25} \textbf{16.346} \\ \cline{1-10}

            &     &      & Euclidean                  & \multirow{12}{*}{4622} & 3200         &\multirow{12}{*}{22.98\%}  & 15.91\%          &  \multirow{12}{*}{21.883}& 109.249 \\
            &     &Elbow & Multi-Step                 &                        & 2760         &                           & 13.72\%          &                          & 103.912 \\
            &     &      &\cellcolor{gray!25} \textbf{Spatio-temporal}   &                        &\cellcolor{gray!25}\textbf{2760} &                           &\cellcolor{gray!25} \textbf{13.72\%} &                          &\cellcolor{gray!25} \textbf{77.111}  \\    
                 \cline{3-4} \cline{6-6}  \cline{8-8} \cline{10-10}
            &     &           & Euclidean             &                        & 3570         &                           & 17.75\%          &                          & 112.909 \\
            &     &Silhouette & Multi-Step            &                        & 2100         &                           & 10.44\%          &                          & 106.304 \\
        ICA &20110&      &\cellcolor{gray!25} \textbf{Spatio-temporal}   &                        &\cellcolor{gray!25} \textbf{2100} &                           &\cellcolor{gray!25} \textbf{10.44\%}  &                          &\cellcolor{gray!25} \textbf{76.271} \\ 
                 \cline{3-4} \cline{6-6}  \cline{8-8} \cline{10-10}
            &     &           & Euclidean             &                        & 4230         &                           & 21.03\%          &                          & 113.167 \\
            &     & Gap       & Multi-Step            &                        & 2300         &                           & 11.44\%          &                          & 107.561 \\
            &     &       &\cellcolor{gray!25} \textbf{Spatio-temporal}  &                        &\cellcolor{gray!25} \textbf{2300}&                           &\cellcolor{gray!25} \textbf{11.44\%} &                          &\cellcolor{gray!25} \textbf{76.991} \\   \cline{3-4} \cline{6-6}  \cline{8-8} \cline{10-10}
            &     &           & Euclidean             &                        & 4534         &                           & 22.55\%          &                          & 113.183 \\
            &     &Canopy     & Multi-Step            &                        & 3492         &                           & 17.36\%          &                          & 108/197 \\
            &     &       &\cellcolor{gray!25} \textbf{Spatio-temporal}  &                        &\cellcolor{gray!25}\textbf{3492} &                           &\cellcolor{gray!25} \textbf{17.36\%} &                          &\cellcolor{gray!25} \textbf{78.431} \\    
        \bottomrule
        \end{tabular}} % Close the font size setting
        }
        \end{table}
\end{small}

\begin{figure}[h]
  \begin{minipage}{0.33\linewidth}
    \includegraphics[width=\linewidth]{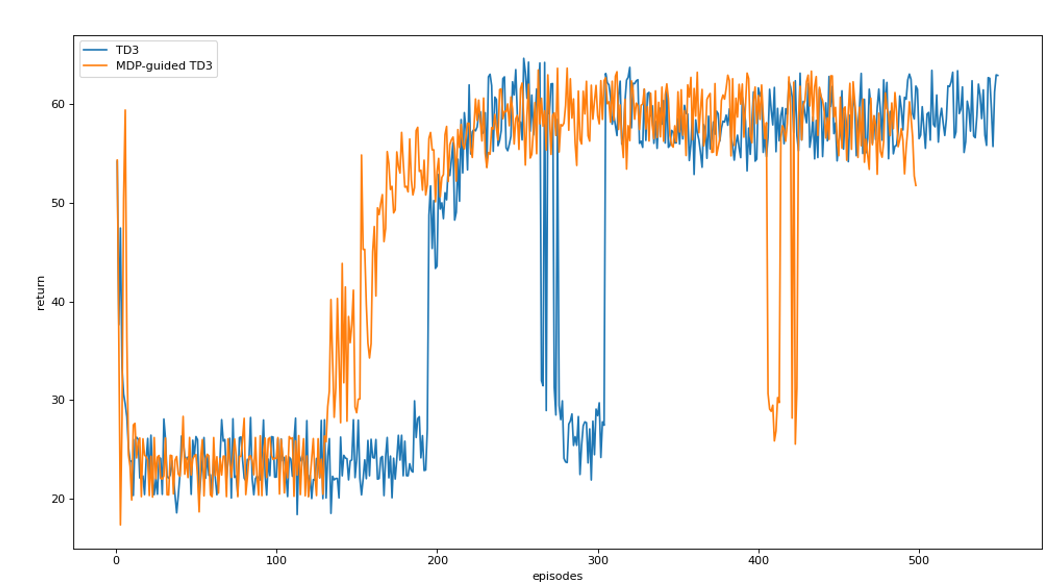} 
    \subcaption{ACC}
    \label{fig:ACC_td3_0.3_0.7}
  \end{minipage}
  \begin{minipage}{0.33\linewidth}
    \includegraphics[width=\linewidth]{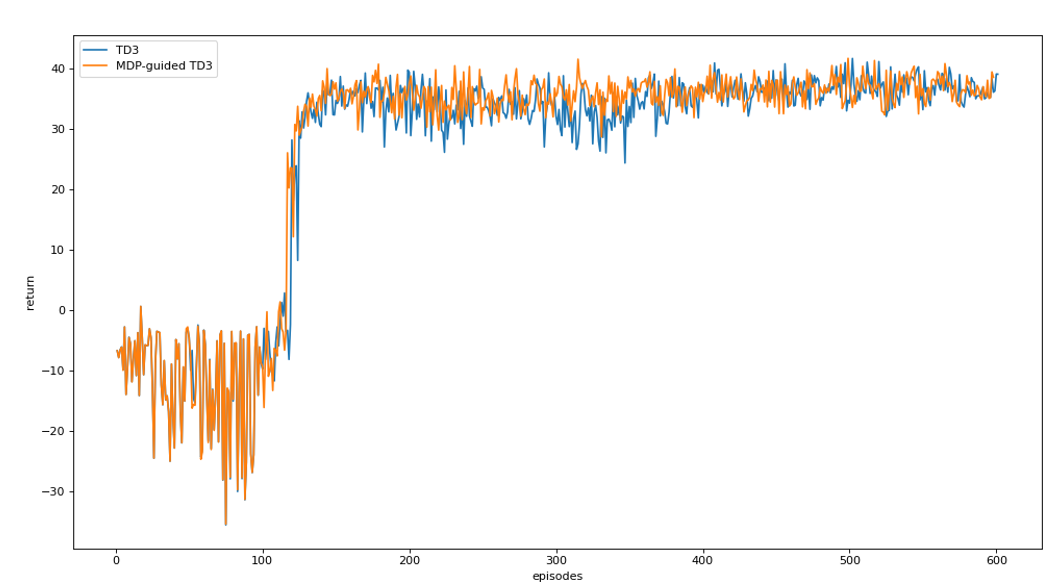}
    \subcaption{LKA}
    \label{fig:LKA_TD3_0.3_0.7}
  \end{minipage}
  \begin{minipage}{0.33\linewidth}
    \includegraphics[width=\linewidth]{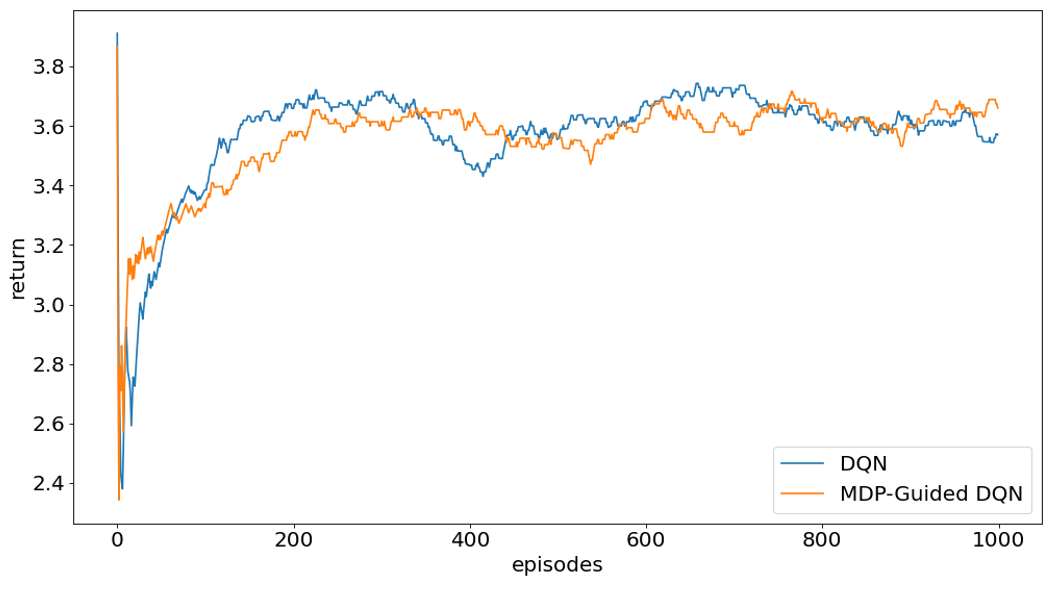}
    \subcaption{ICA}
    \label{fig:ICA_TD3_0.5_0.5}
  \end{minipage}
  \caption{Abstract MDP-Guided DRL for Various Scenarios}
  \label{fig:MDP-Guided_DRL}
\end{figure}

\textbf{RQ3: Abstract MDP-Guided Learning Process} Fig.~\ref{fig:MDP-Guided_DRL} features three line graphs, each corresponding to distinct scenarios in RL: (a) ACC, (b) LKA, and (c) ICA. Each graph portrays the performance of two models across numerous episodes: the conventional model (denoted as TD3 or DDPG) and the abstract MDP-Guided model (referred to as MDP-guided TD3 or MDP-guided DQN). The vertical axis represents the obtained reward, while the horizontal axis indicates the episode count.

An examination of Fig.~\ref{fig:MDP-Guided_DRL} underscores that, across all considered scenarios, the abstract MDP-Guided models (depicted in orange) consistently surpass the performance of the traditional models (depicted in blue). Of particular significance is the discernibly accelerated escalation in reward exhibited by the abstract MDP-Guided models, indicative of more efficient learning dynamics. This acceleration is particularly pronounced in the initial episodes, where the abstract MDP-Guided models attain higher rewards at a faster pace in comparison to their traditional counterparts. Moreover, our proposed approach demonstrates superior guidance, especially in intricate scenarios.

The discerned trends across all three scenarios affirm the pivotal role of abstract models in significantly enhancing the efficiency of the RL process. By adeptly simplifying the inherent complexity of the environment and channeling learning efforts towards critical aspects, abstract models expedite the policy learning process. This, in turn, results in a more expeditious and effective training regimen, thereby effectively addressing the posed research question (RQ3).
% 比较对象加参考文献
\begin{small}
    \begin{table}[htb]
        \centering
        \caption{Semantic Gap Analysis of Abstract MDP with PRISM}
        \label{tab:Semantic_error}
        \resizebox{\textwidth}{!}{%=
        {\fontsize{8}{10}\selectfont 
        \begin{tabular}{c|c|c|c|c|c}
        \toprule
        \scriptsize Case Study                  & \scriptsize Metric & \scriptsize Properties        &  \scriptsize Verification \scriptsize Result & \scriptsize Real Value & \scriptsize Error \\ 
        \midrule
        \multirow{6}{*}{\scriptsize LKA} & \multirow{2}{*}{\scriptsize Euclidean}          &\scriptsize  Rmin=?{[}C\textless{}=51{]}               & 44.43       & 48.50 & 4.07 \\
        &                                                                                  & \scriptsize Pmax=?{[}F\textless{}=51;isOutOfLane=1{]} & 0.13\%      & 0.00\% & 0.13 \\ \cline{2-6} 
                                          & \multirow{2}{*}{\scriptsize Multi-Step} & \scriptsize Rmin=?{[}C\textless{}=51{]}                    & 44.72       & 48.50   & 3.78 \\
        &                                                                           & \scriptsize Pmax=?{[}F\textless{}=51;isOutOfLane=1{]}      & 0.13\%      & 0.0\% & 0.13 \\ \cline{2-6} 
                                          &\cellcolor{gray!25} &\cellcolor{gray!25} \scriptsize \textbf{Rmin=?{[}C\textless{}=51{]}}    &\cellcolor{gray!25} \textbf{46.92} &\cellcolor{gray!25} \textbf{48.50} &\cellcolor{gray!25} \textbf{1.58} \\ 
        &\multirow{-2}{*}{\cellcolor{gray!25}\scriptsize \textbf{Spatio-temporal Value}}                                               & \cellcolor{gray!25}\scriptsize  \textbf{Pmax=?{[}F\textless{}=51;isOutOfLane=1{]}}     & \cellcolor{gray!25}\textbf{0.10\%} & \cellcolor{gray!25}\textbf{0.00\% }&\cellcolor{gray!25} \textbf{0.10} \\ \hline
        \multirow{9}{*}{\scriptsize ACC} & \multirow{3}{*}{\scriptsize Euclidean}   & \scriptsize Rmin=?{[}C\textless{}=51{]}               & 57.53 & 59.94 & 2.41 \\
        &                                                                      & \scriptsize Pmax=?{[}F\textless{}=51;isOutOfLane=1{]}      & 0.7\% & 0.00\% & 0.7 \\ 
        &                                                                      & \scriptsize Pmax=?{[}F\textless{}=51;isCrashed=1{]}        & 0.06\% & 0.00\% & 0.06 \\ \cline{2-6} 
                                         & \multirow{3}{*}{\scriptsize Multi-Step}   & \scriptsize Rmin=?{[}C\textless{}=51{]}              & 55.98 & 59.94 & 3.96 \\
        &                                                                      & \scriptsize Pmax=?{[}F\textless{}=51;isOutOfLane=1{]}      & 1.00\%    & 0.00\% & 1.00 \\ 
        &                                                                      & \scriptsize Pmax=?{[}F\textless{}=51;isCrashed=1{]}        & 0.06\% & 0.00\% & 0.06 \\ \cline{2-6} 
                                         &\cellcolor{gray!25} &\cellcolor{gray!25} \scriptsize \textbf{Rmin=?{[}C\textless{}=51{]}}     &\cellcolor{gray!25} \textbf{60.33} &\cellcolor{gray!25} \textbf{59.94} &\cellcolor{gray!25} \textbf{-0.39} \\ 
        & \cellcolor{gray!25}                                                                     &\cellcolor{gray!25} \scriptsize  \textbf{Pmax=?{[}F\textless{}=51;isOutOfLane=1{]}}     &\cellcolor{gray!25} \textbf{0.01\%} &\cellcolor{gray!25} \textbf{0.00\%} &\cellcolor{gray!25} \textbf{0.01} \\ 
        &  \multirow{-3}{*}{ \cellcolor{gray!25} \scriptsize \textbf{Spatio-temporal Value}}                                                                     &\cellcolor{gray!25} \scriptsize \textbf{Pmax=?{[}F\textless{}=51;isCrashed=1{]}}        &\cellcolor{gray!25} \textbf{0.19\%} &\cellcolor{gray!25}\cellcolor{gray!25} \textbf{0.00\%} &\cellcolor{gray!25} \textbf{0.19} \\ \hline
        \multirow{8}{*}{\scriptsize ICA} & \multirow{3}{*}{\scriptsize Euclidean}   & \scriptsize Rmin=?{[}C\textless{}=60{]}          & 8.38 & 9.36 & 0.98 \\
        &                                                                      & \scriptsize Pmax=?{[}F\textless{}=60;isCrashed=1{]}   & 18.73\% & 20.80\% & 2.07 \\ 
        &                                                                      & \scriptsize Pmax=?{[}F\textless{}=60;reachDest=1{]}   & 0.17\% & 4.60\% & 4.43 \\ \cline{2-6} 
                                         &  \multirow{3}{*}{\scriptsize Multi-Step}   & \scriptsize Rmin=?{[}C\textless{}=60{]}        & 8.99 & 9.36 & 0.37 \\
        &                                                                      & \scriptsize Pmax=?{[}F\textless{}=60;isCrashed=1{]}   & 17.33\% & 20.80\% & 3.47 \\ 
        &                                                                      & \scriptsize Pmax=?{[}F\textless{}=60;reachDest=1{]}   & 3.25\% & 4.60\% & 1.35 \\ \cline{2-6} 
                                         &\cellcolor{gray!25}& \cellcolor{gray!25} \scriptsize \textbf{Rmin=?{[}C\textless{}=60{]}}&\cellcolor{gray!25} \textbf{9.38} &\cellcolor{gray!25} \textbf{9.36} & \cellcolor{gray!25} \textbf{ -0.02} \\ 
        & \cellcolor{gray!25}                                                                     &\cellcolor{gray!25} \scriptsize \textbf{Pmax=?{[}F\textless{}=60;isCrashed=1{]}}   &\cellcolor{gray!25} \textbf{19.35\%} &\cellcolor{gray!25} \textbf{20.80\%} &\cellcolor{gray!25} \textbf{1.45} \\ 
        & \multirow{-3}{*}{\cellcolor{gray!25}  \scriptsize \textbf{Spatio-temporal Value}}                                                                  &\cellcolor{gray!25} \scriptsize \textbf{Pmax=?{[}F\textless{}=60;reachDest=1{]}}   &\cellcolor{gray!25} \textbf{4.50\%} &\cellcolor{gray!25} \textbf{4.60\%} &\cellcolor{gray!25} \textbf{0.10} \\
        \bottomrule
        \end{tabular}} % Close the font size setting
        }
        \end{table}
\end{small}
% Prism 验证地方写明确 为什么要用Prism 以及性质如何确定

\textbf{Response to RQ3:} Fig.~\ref{fig:MDP-Guided_DRL} strongly supports the pivotal role of abstract models in guiding the DRL process. Offering a structured approach to the state space and incorporating domain knowledge through abstraction, these models enhance data utilization and generalization. This leads to accelerated convergence towards higher rewards, signifying a more rapid training process. Furthermore, the improved performance in initial episodes suggests that abstract models effectively address challenges related to data scarcity, leveraging abstracted information to guide the learning algorithm toward profitable strategies early in the training process.

\begin{small}
    \begin{figure}[h]
  \begin{minipage}{0.5\linewidth}
    \includegraphics[width=\linewidth, height=2in]{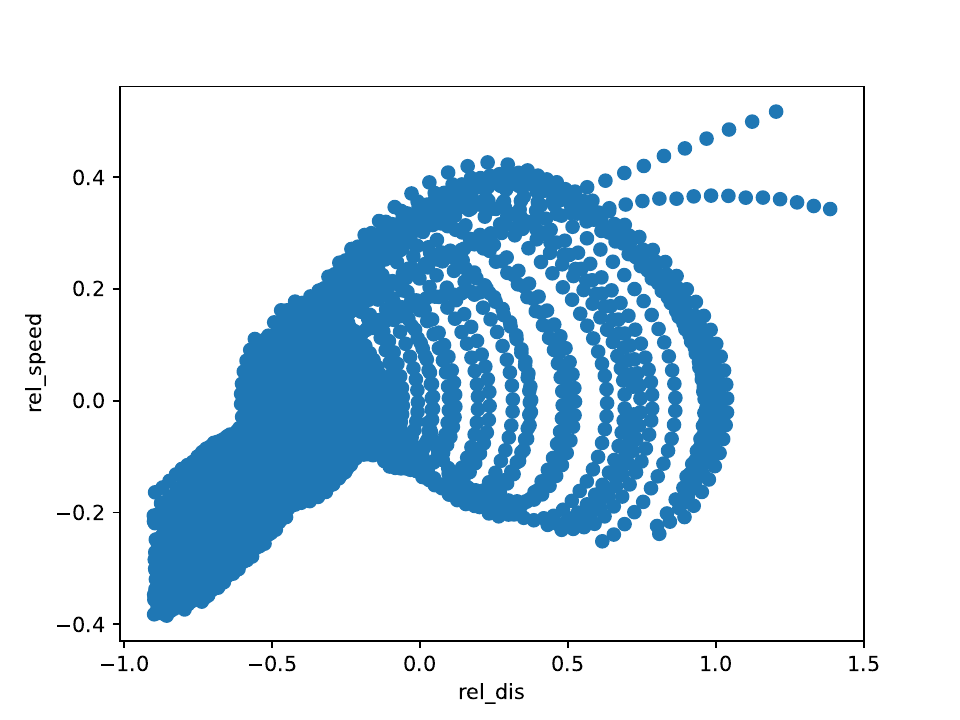} 
    \subcaption{ACC Concrete Traces}
    \label{fig:ACC_concrete_Traces}
  \end{minipage}
  \begin{minipage}{0.5\linewidth}
    \includegraphics[width=\linewidth, height=2in]{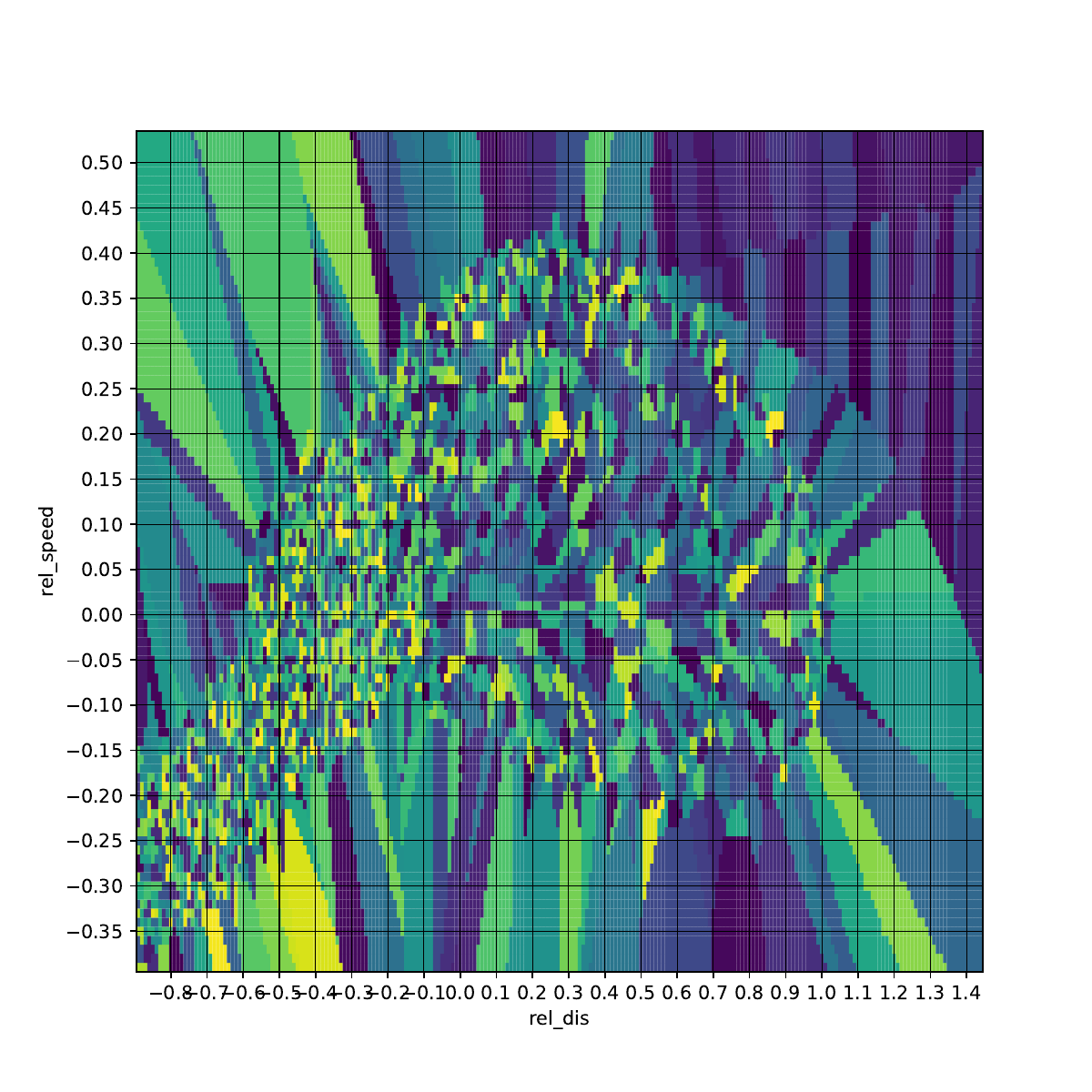} % 调整height的值
    \subcaption{ACC Abstract Traces}
    \label{fig:ACC_Abstract_Traces}
  \end{minipage}
  \caption{From Concrete Traces to Abstract Traces}
  \label{fig:trace_distribution}
  % \centering
  \small
    Using the ACC scenario as an illustration, the horizontal axes in both Fig.~\ref{fig:ACC_concrete_Traces} and Fig.~\ref{fig:ACC_Abstract_Traces} designate relative speed, whereas the vertical axes signify relative distance, both subjected to normalization. Fig.~\ref{fig:ACC_concrete_Traces} delineates the unprocessed traces of the ADS. Conversely, Fig.~\ref{fig:ACC_Abstract_Traces} elucidates abstract traces, encapsulating data that transcends the boundaries of the exploration scope. Distinct hues correspond to disparate abstract states in the representation.
\end{figure}
\end{small}

%但是这些方法都不能进行验证
\section{Related Work}\label{Section:5}
\textit{\textbf{Action and State Abstraction in MDPs.}}
The innovative concept of MDP action abstraction proves to be a strategic solution for alleviating computational burdens and enhancing problem-solving efficiency by compressing the action space while maintaining solution quality. Chen and Xu~\cite{chen2016statistical} pioneered a method grounded in discrete Fourier transform for action abstraction in MDPs with deterministic uncertainty. Extending this paradigm to MDPs with continuous action spaces, Omidshafiei~\cite{omidshafiei2015decentralized} applied the Fourier transform. Additionally, Bita~\cite{ijcai2023p347} contributes a comprehensive framework that leverages the nondeterministic situation calculus and ConGolog programming language to abstract agent actions in nondeterministic domains, facilitating strategic reasoning and synthesis.

Simultaneously, when employing function approximation to abstract states, involving the intricate process of mapping the concrete space to a lower-dimensional counterpart optimized through RL objectives. Studies integrating NNs and Hierarchical Reinforcement Learning (HRL), such as feudal HRL~\cite{andreas2017modular,oh2017zero} and option-critic with NNs~\cite{jiang2015abstraction,guo2021state}, actively address the nuanced realms of state and temporal abstraction. Abel~\cite{abel2018state} introduces transitive and PAC state abstractions, triumphing in sample complexity reduction despite potential performance drawbacks. Misra~\cite{misra2020kinematic} introduces HOMER, a pioneering sample-efficient exploration and DRL algorithm tailored for rich observation environments, guaranteeing provable efficiency and computational effectiveness in specific scenarios. However, these methods, while effective in specific scenarios, may lack semantic preservation in the process of abstraction.

%相关工作 别人使用解决什么问题 我们怎么怎么做～
\noindent \textit{\textbf{Abstract MDPs for DRL}}. The realm of MDP abstraction, meticulously preserving transition and reward structures~\cite{lavaei2022constructing}, emerges as a linchpin for augmenting RL efficiency and generalization. Existing options, such as option-bisimulation~\cite{castro2011planning,castro2010using}, grapple with computational intricacies. Abel et al.~\cite{abel2020value} pioneer state-abstraction-option classes with insightful suboptimality bounds. Vans~\cite{van2020mdp} introduces MDP homomorphic networks, harnessing symmetries for enhanced convergence. Guo~\cite{guo2021state}  with a Multi-Step metric for state-temporal abstraction.
Junges~\cite{junges2022abstraction} takes advantage of the inherent hierarchy of the Markov decision process and divides the state space into macro-level and sub-level. It regards the unresolved sub-level as an uncertainty for constraint and progressive analysis, to reduce the state space explosion problem.
Feng~\cite{feng2023dense} unfolds the potential of edited MDPs, efficiently learning safety-critical states from naturalistic driving data, thus showcasing accelerated testing and training of safety-critical autonomous systems. The existing work on abstract MDPs demonstrates notable contributions but faces challenges in terms of verification, safety, and generalization across diverse environments. Strengthening these aspects is essential for advancing the reliability and applicability of abstraction techniques in real-world scenarios.  

\section{Conclusion and Future Work}\label{Section:6}
Our approach to abstract modeling using spatio-temporal value semantics represents a substantial leap in ensuring the dependability of machine learning systems, particularly in decision-making processes for ICPS employing DRL. This method is characterized by its universal application across a variety of ICPS scenarios, demonstrating its robustness and versatility. However, it's crucial to recognize certain limitations, such as the learning process efficiency for the abstract model, which we aim to enhance in our future work. 

Expanding upon this, the next phase of our research will focus on refining the learning algorithms to accelerate the training phase without compromising the model's integrity. Additionally, moving beyond experimental validations, we plan to incorporate formal theorem-based evaluations to establish the equivalence between our abstract models and their respective true MDPs. This shift towards a more rigorous theoretical framework, such as bisimulation, will allow for a deeper understanding and validation of the models' accuracy and reliability. 

Moreover, our future endeavors aim to broaden the application of our abstraction technique to more complex ICPS domains. This extension includes optimizing the efficiency of state exploration in DRL training, a crucial aspect for effective navigation in intricate and dynamic environments. Additionally, we intend to explore reachability and robustness aspects within these systems, ensuring that our models not only make predictions but also respond adaptively to real-world scenarios. Through these comprehensive research efforts, our overarching objective is to make significant contributions to the field of abstract modeling, enhancing its effectiveness and reliability in safety-critical and dynamic environments. Addressing these challenges, ensuring model safety, and establishing a cohesive framework remain pivotal for the successful and robust application of abstraction techniques in practical scenarios.

\newpage
\bibliographystyle{splncs04}
\bibliography{reference}

\end{document}